\documentclass[runningheads]{llncs}

\usepackage{eccv}

\usepackage{eccvabbrv}

\usepackage{graphicx}
\usepackage{booktabs}

\usepackage[accsupp]{axessibility}  

\usepackage[pagebackref,breaklinks,colorlinks,citecolor=eccvblue]{hyperref}

\usepackage{orcidlink}
\usepackage{algorithm}
\usepackage{algorithmicx}
\usepackage{algpseudocode}  
\usepackage{tablefootnote}
\begin{document}
	
	\title{Deep Contrastive Multi-view Clustering under Semantic Feature Guidance} 
	
	
	\author{Siwen Liu\inst{1}\and
		Jinyan Liu\inst{1}\and
		Hanning Yuan\inst{1}\and
		Qi Li\inst{1}\and
		Jing Geng\inst{1}\and
		Ziqiang Yuan\inst{1}\and
		Huaxu Han\inst{1}
		}
	
	\authorrunning{Liu et al.}
	
	\institute{Beijing Institute of technology, China \\}
	
	\maketitle

	\begin{abstract}
	Contrastive learning has achieved promising performance in the field of multi-view clustering recently. However, the positive and negative sample construction mechanisms ignoring semantic consistency lead to false negative pairs, limiting the performance of existing algorithms from further improvement. To solve this problem, we propose a multi-view clustering framework named \underline{D}eep \underline{C}ontrastive \underline{M}ulti-view \underline{C}lustering under \underline{S}emantic feature guidance (DCMCS) to alleviate the influence of false negative pairs. Specifically, view-specific features are firstly extracted from raw features and fused to obtain fusion view features according to view importance. To mitigate the interference of view-private information, specific view and fusion view semantic features are learned by cluster-level contrastive learning and concatenated to measure the semantic similarity of instances. By minimizing instance-level contrastive loss weighted by semantic similarity, DCMCS adaptively weakens contrastive leaning between false negative pairs. Experimental results on several public datasets demonstrate the proposed framework outperforms the state-of-the-art methods.
	\end{abstract}

	\section{Introduction}
	\label{sec:intro}
	Multi-view clustering(MVC) has gained interest in the last few years and has been widely applied in the field of  biology\cite{zhang2023multi}, medicine \cite{chen2023deep, yang2023exploring}, social network \cite{yu2018web}, agriculture \cite{ramon2018multi}, and so forth. Traditional methods for MVC include non-negative matrix factorization\cite{yin2022incomplete, zong2018multi}, latent representation learning\cite{jin2021model,xie2020adaptive}, graph learning \cite{kang2020multi, wang2021consistent}, and tensor learning \cite{chen2015total, liu2013multi}. Many of the traditional MVC methods suffer from poor representation and high computational complexity, resulting in limited performance in complex scenarios with real data\cite{guo2019anchors}. Deep learning develops superior features with its strong representation ability, so deep MVC is becoming popular. Deep MVC methods can be categorized into four subgroups: subspace clustering\cite{lan2024double, peng2020deep}, graph-based deep clustering \cite{tan2023sample, wen2023highly}, deep representation clustering \cite{trosten2021reconsidering,yang2021deep}, and spectral clustering\cite{chen2023multiview, liang2020multi}. Compared to other approaches, deep representation clustering often employs a simple encoder-decoder structure and may be more flexible\cite{fang2023comprehensive}.
	
	Effective deep MVC depends on learning discriminative common information from multi-view data. Recently, contrastive learning has been integrated into deep MVC because of its ability to capture high-level semantics while discarding irrelevant information\cite{xu2024self}. Ke \etal~\cite{ke2021conan, ke2022mori} design a fusion network to extract common information. Xu \etal~\cite{xu2022multi} carry out distinct goals at different feature levels to resolve the conflict between consistency and reconstruction objectives. Hu \etal~\cite{hu2023joint} establish triple features through contrastive learning using feature-oriented alignment, commonality-oriented, and cluster-level consistency. Even though these contrastive learning-based deep MVC methods have achieved  important progress, there are still some issues that need to be resolved: (1) For directly applying contrastive loss such as InfoNCE\cite{chen2020simple}, the performance of many methods(\eg\cite{hu2023joint, xu2022multi}) is impaired by false negative pairs. Minimizing the contrastive loss may increase the feature dissimilarity of instances sharing the same cluster label due to false negative pairs, which contradicts with clustering objective and leads to unfriendly representation learning for clustering. (2)Some methods(\eg\cite{ke2021conan,ke2022mori}) fuse specific views to obtain a fusion view containing common information. However, the view-private information is inevitably introduced to fusion view features and transferred to contrastive learning interfering with the quality of clustering. 
	
	In this paper, we propose a novel semantics-guided multi-view contrastive clustering framework to address the above issues. Our framework, as illustrated in \cref{fig:stru}, consists of a view-specific feature fusion module and a cross-view double-level contrastive module. In the view-specific feature fusion module, auto-encoders are used to generate view-specific features, and weighted fusion is employed to generate the fusion view features. The cross-view double-level contrastive fusion module employs two consistency objectives: semantics-guided instance-level contrastive learning and cluster-level contrastive learning. Instance pair weights measured by semantic features are applied to mitigate the impact of false negative pairs. Furthermore, in order to reduce the influence of view-private information, common information are concentrated when calculating instance pair weights. Compared with previous work, our contributions are as follows:
	\renewcommand{\labelitemi}{\textbullet}
	\begin{itemize}
		\item A DCMCS framework is proposed to lessen the impact of false negative pairs in instance-level contrastive learning. The objective of contrastive learning is made consistent with the clustering objective by using instance pair weights obtained from semantic features to learn cluster-friendly features.
		\item When calculating the instance pair weights, the common information of the fusion view are concentrated in DCMCS. It relieves the interference of the view-private information in the fusion view to achieve better clustering results.
		\item In the experiment, we show how effectively the instance pair weights and the fusion view's common information work. Experimental results on several public datasets demonstrate our framework outperforms the state-of-the-art methods.
	\end{itemize}
	
		The rest of the paper is organized as follows. The relevant work is presented in \cref{sec:relate}. The proposed framework is presented in \cref{sec:method}. Comprehensive experiments and findings are reported in \cref{sec:exper}. \cref{sec:con} provides the conclusion and future work.

	\section{Related Work}
	\label{sec:relate}
	\subsection{Multi-view clustering.}
	With the development of deep learning, deep MVC has an important position in MVC methods. The four categories of deep MVC methods are deep representation learning\cite{chen2023incomplete, li2019deep,trosten2021reconsidering, yang2021deep},
	deep graph learning\cite{li2020bipartite, tan2023sample, wen2023highly},
	subspace clustering\cite{lan2024double, long2023multi, peng2020deep},
	and spectral clustering\cite{chen2023multiview, liang2020multi, zhong2023self}; the latter three are frequently combined\cite{cui2023deep, li2023deep, yang2023one}.
	In deep representation learning, contrastive methods\cite{hu2023joint, trosten2021reconsidering, xu2022multi},
	collaborative methods\cite{yang2021deep, zhou2024mcoco}, adversarial methods\cite{li2019deep, zhou2020end}, and distillation methods\cite{chen2023incomplete, liu2022inconsistency} can be used to acquire common information. Yang \etal~\cite{yang2021deep} optimize view features using intra-view collaborative learning and gain complementing information through inter-view collaborative learning. Li \etal~\cite{li2019deep} employ adversarial learning to further capture data distribution. A distillation approach is presented by Chen \etal~\cite{chen2023incomplete} that uses a teacher model with complete views to direct a student model with lacking views. Contrastive learning is frequently employed in deep MVC since it supports the clustering objective. We present a semantics-guided contrastive multi-view clustering framework that takes advantage of the consistency and complementarity of views.
	
	\subsection{Contrastive learning.}
	Contrastive learning\cite{chen2020simple, he2020momentum} has proven to be quite effective in self-supervised learning in recent years. By creating positive and negative pairs, contrastive learning is able to obtain discriminative features. New perspectives on the difficulties of deep clustering are offered by the development of contrastive clustering\cite{li2021contrastive}. In general, there are two types of multi-view contrastive clustering: instance-level contrastive learning
	\cite{trosten2021reconsidering, yan2023gcfagg} and cluster-level contrastive learning\cite{chen2023deepc}. \cite{hu2023joint, xu2022multi} make use of both. False negative pairs occur when setting negative pairs in the popular instance-level contrastive learning technique. In order to mitigate this issue, Trosten \etal~\cite{trosten2021reconsidering} use negative pairs consisting of instances assigned to distinct clusters, whereas Yan \etal~\cite{yan2023gcfagg} add structure relationship to  negative pairs. Semantics-guided instance-level contrastive learning and cluster-level contrastive learning are utilized in our framework. To reduce false negative pairs, instance pair weights derived from semantic features are introduced.
	
	\section{Method}
	\label{sec:method}
	In this section, we first explain the definitions used and the goal to be achieved in the framework. Next, we introduce the architecture of the framework. Finally, the total loss and entire optimization procedure are shown.
	\begin{figure}[t]
		\centering
		\includegraphics[height=5.5cm,width=12cm]{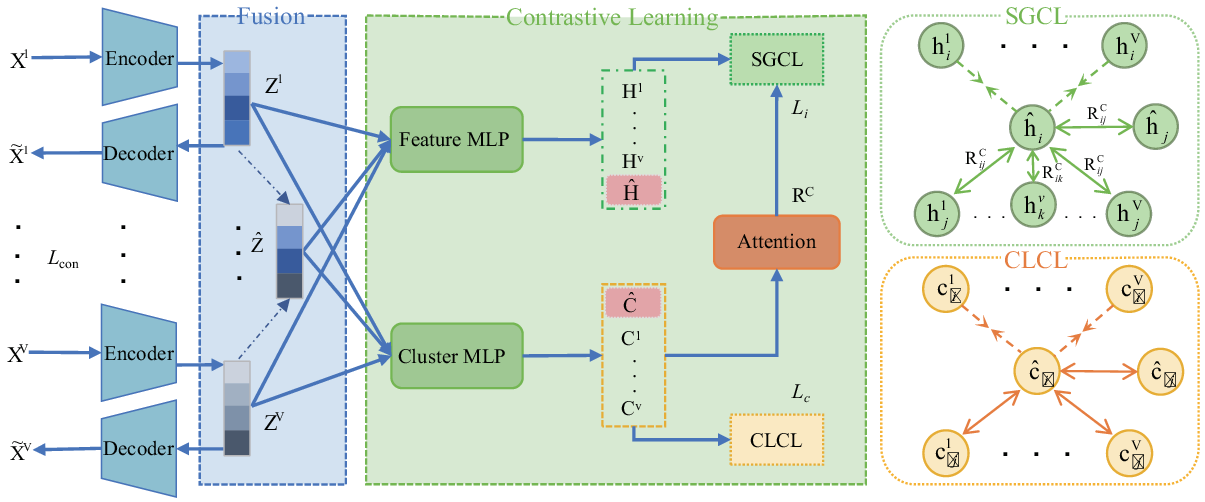}
		\caption{The framework of DCMCS. The view-specific features $\mathrm{Z}^{v}$, the fusion view features
		$\hat{\mathrm{Z}}$, the instance-level features $\mathrm{H}^{v}$,$\hat{\mathrm{H}}$ and the semantic features $\mathrm{C}^{v}$,$\hat{\mathrm{C}}$ are learned from the raw features $\mathrm{X}^{v}$. The reconstruction objective $L_{\mathrm{con}}$ is
		individually conducted on $\mathrm{Z}^{v}$. In semantics-guided contrastive learning(SGCL) and cluster-level contrastive learning(CLCL) modules,
		two contrastive losses (i.e., $L_i$ and $L_c$) are conducted on the instance-level features and cluster-level features, respectively.  Moreover, $\mathrm{R^C}$ represents the weight matrix generated from semantic features to establish the relationship between negative pairs in $L_i$.}
		\label{fig:stru}
	\end{figure}
	\subsection{Proposed Statement}
	Given a multi-view dataset $\chi=\left\{\mathrm{X}^{v} \in \mathbb{R}^{\mathrm{N} \times \mathrm{D}_{v}}\right\}_{v=1}^{\mathrm{V}}$ with V views and N samples,  $\mathrm{X}^{v}=[\mathrm{x}_{1}^{v},\mathrm{x}_{2}^{v},\ldots,\mathrm{x}_{\mathrm{N}}^{v}]$ denotes the instances of the \textit{v}-th view. $\mathrm{D}_{v}$ denotes the feature dimension of the \textit{v}-th  view. Considering that there are K clusters in the dataset, the N samples are ultimately divided into K clusters according to their similarity.

	\subsection{View-specific feature fusion}
	\label{sec:line-numbering}
	The view-specific feature fusion module uses several view-specific autoencoders, as shown in \cref{fig:stru}, to extract the features of each view. 
	In particular, we designate the encoder $f_{\theta_{v}}^v$ and the decoder $g_{\eta_v}^v$ for the \textit{v}-th view, where $\mathrm{\theta}_{v}$ and  $\mathrm{\eta}_{v}$ are parameters of the encoder and decoder respectively. Encoder $f_{\theta_{v}}^v$ projects raw features into view-specific feature space by
	\begin{equation}
		\mathrm{z}_{i}^{v}=f_{\theta^{v}}^{v}\left(\mathrm{x}_{i}^{v}\right)
	\end{equation}
	
	where $\mathrm{Z}^{v}=[\mathrm{z}_{1}^{v},\mathrm{z}_{2}^{v},\ldots,\mathrm{z}_{\mathrm{N}}^{v}]$, $\mathrm{Z}^{v} \in \mathbb{R}^{\mathrm{N} \times \mathrm{d}}$. $\mathrm{d}$ is the dimension of the view-specific features. Inspired by \cite{guo2017improved}, DCMCS employ decoder $g_{\eta_v}^v$ to reconstruct $\mathrm{z}_{i}^{v}$ to learn sufficient discriminative information avoiding the model collapsing.
	\begin{equation}
		\widetilde{\mathrm{x}}_i^{v}=g_{\eta^{v}}^{v}(\mathrm{z}_i^{v})=g_{\eta^{v}}^{v}(f_{\theta^{v}}^{v}(\mathrm{x}_i^{v}))
	\end{equation}
	
	The reconstruction loss for all views is defined as:
		\begin{equation}
		L_{\mathrm{con}}=\sum_{{v}=1}^{V}\left\|\mathrm{X}^{{v}}-\widetilde{\mathrm{X}}^{{v}}\right\|_{2}^{2}=\sum_{{v}=1}^{V}\sum_{i=1}^{N}\left\|\mathrm{x}_{i}^{{v}}-g_{\eta^{v}}^{{v}}\left(f_{\theta^{v}}^{{v}}\left(\mathrm{x}_{i}^{{v}}\right)\right\|_{2}^{2}\right. 
		\label{eq:con}
	\end{equation}

	Weighted fusion is a simple and effective method for obtaining more discriminative fusion view features by leveraging the consistency and complementarity of multi-view data. Considering the importance of different views, we fuse the views using adaptive weighting to obtain the fusion view features $\mathrm{\hat{Z}}\in\mathbb{R}^{\mathrm{N}\times\mathrm{d}}$, where $\mathrm{\hat{Z}}=[\mathrm{\hat{z}}_{1},\mathrm{\hat{z}}_{2},\ldots,\mathrm{\hat{z}}_{\mathrm{N}}]$. $\mathrm{\hat{z}}_{i}$ is defined as:
	\begin{equation}
		\hat{\mathrm{z}}_i=\sum_{{v}=1}^Vw_{v}\mathrm{z}_i^{v} 
		\label{eq:fusion}
	\end{equation}
	
	where $w_{v}$ is the \textit{v}-th view's weight, $\sum_{{v}=1}^{{V}}w_{{v}}=1$. The weights are obtained by loss function optimization, reflecting the importance of different views. 
	
	\subsection{Cluster-level contrastive learning}
	Contrastive learning(CL) effectively captures high-level semantics while removing irrelevant information since it directly increases the feature similarity between semantically relevant instances. We apply cluster-level contrastive learning to improve cluster consistency across multiple views. A two-layer linear MLP with parameter $\mathrm{W}_\mathrm{C}$ is utilized to map the view-specific features and fusion view features to K-dimension space. K is the number of clusters. A Softmax layer is attached to obtain the semantic features of each view $\mathrm{C}^v=[\mathrm{c}_1^v,\mathrm{c}_2^v,\ldots,\mathrm{c}_{\mathrm{N}}^v]$, $\mathrm{C}^v\in\mathbb{R}^{\mathrm{N\times K}}$ and the semantic features of the fusion view $ \mathrm{\hat{C}}=[\mathrm{\hat{c}}_{1},\mathrm{\hat{c}}_{2},\ldots,\mathrm{\hat{c}}_{\mathrm{N}}]$, $\widehat{\mathrm{C}}\in\mathbb{R}^{\mathrm{N}\times\mathrm{K}}$. The semantic features are described by the probability that the instance belongs to each cluster. The instance's cluster label corresponds to the cluster with the highest probability. Cluster-level feature $\mathrm{C}_{\cdot j}^{v}$ is represented as the probability that each instance belongs to cluster \textit{j} in the \textit{v}-th view.
	
	Since the instances in multiple views that correspond to a single sample share semantic information in common, the instances' cluster assignment probabilities across various views ought to be similar, and cluster-level features from the same cluster should be similar. Cluster-level contrastive learning increases the distance between cluster pairs corresponding to distinct clusters while decreasing the distance between cluster pairs corresponding to the same cluster. For each fusion view cluster-level feature $\hat{\mathrm{c}}_{\cdot j}$, $\left\{\hat{\mathrm{c}}_{\cdot j},\mathrm{c}_{\cdot j}^{v} \right\}_{v=1,...,\mathrm{V}}$ are positive pairs and the rest cluster pairs of $\hat{\mathrm{c}}_{\cdot j}$ are negative pairs.
	
	Cosine similarity is used to quantify the similarity between two clusters, which is explained as follows:
	\begin{equation}
		S\left(\hat{\mathrm{c}}_{\cdot j},\mathrm{c}_{\cdot j}^{v}\right)=\frac{\left\langle\hat{\mathrm{c}}_{\cdot j},\mathrm{c}_{\cdot j}^{v}\right\rangle}{\left\|\hat{\mathrm{c}}_{\cdot j}\right\|\left\|\mathrm{c}_{\cdot j}^{v}\right\|}
	\end{equation}
	
	where $\langle\cdot,\cdot\rangle $ is the dot product operator. The cluster-level contrastive loss is defined as:
	\begin{equation}
		L_{c}=-\frac1{2K}\sum_{j=1}^{K}\sum_{{v}=1}^{V}\log\frac{e^{S(\hat{\mathrm{c}}_{\cdot j},\mathrm{c}_{\cdot j}^v)/\tau_1}}{\sum_{k=1}^{K}e^{S(\hat{\mathrm{c}}_{\cdot j},\mathrm{c}_{\cdot k}^v)/\tau_1}-e^{1/\tau_1}}-\mathrm{H(C)}
	\end{equation}
	
	where $\tau_{1}$ denotes the temperature coefficient and H(C) denotes the entropy of the clustering result. The presence of entropy prevents from falling into a trivial solution\cite{hu2017learning}. H(C) is defined as follows:
	\begin{equation}
		\mathrm{H(C)}=-\sum_{j=1}^K[\sum_{{v}=1}^VP(\mathrm{c}_j^{v})\log P(\mathrm{c}_j^{v})+P(\hat{\mathrm{c}}_j)\log P(\hat{\mathrm{c}}_j)]
	\end{equation}
	
	where $P(\mathrm{c}_{j}^{v})=(\sum_{i=1}^{N}\mathrm{c}_{ij}^{v})/\mathrm{N}$.
	\subsection{Semantics-guided instance-level contrastive learning}
	 The quality of the positive and negative pairs construction are crucial factors in determining the performance of the contrastive MVC methods. A popular contrastive loss function in MVC is InfoNCE\cite{chen2020simple}.
	\begin{equation}
		L_{\mathrm{InfoNCE}}=-\log{(\sum_{i=1}^{N}\frac{e^{S(z_i,z_i^+)/\tau}}{\sum_{j=1}^Ne^{S(z_i,z_j)/\tau}})}
	\end{equation}
	
	InfoNCE treats the instances in the different views corresponding to an individual sample as positive pairs and directly regards all other non-positive instances as negative pairs. This easily brings false negative pairs problems in that instances sharing the same cluster label are constructed as negative pairs, leading to cluster-unfriendly representation learning. To solve this problem, we propose a semantics-guided instance-level contrastive method by instance pair weights.
	
	When calculating instance pair weights,  rather than concentrating solely on the information in the specific views, we pay attention to the common information of the fusion view. The consistency and complementarity in the fusion view semantic features and specific view semantic features are taken into account.  It can generate weights that are more precise and lessen the impact of view-private information in the fusion view. It produces superior clustering results, as demonstrated in \cref{sec:ablation}.
	
	\begin{figure}[t]
		\centering
		\includegraphics[height=4.5cm]{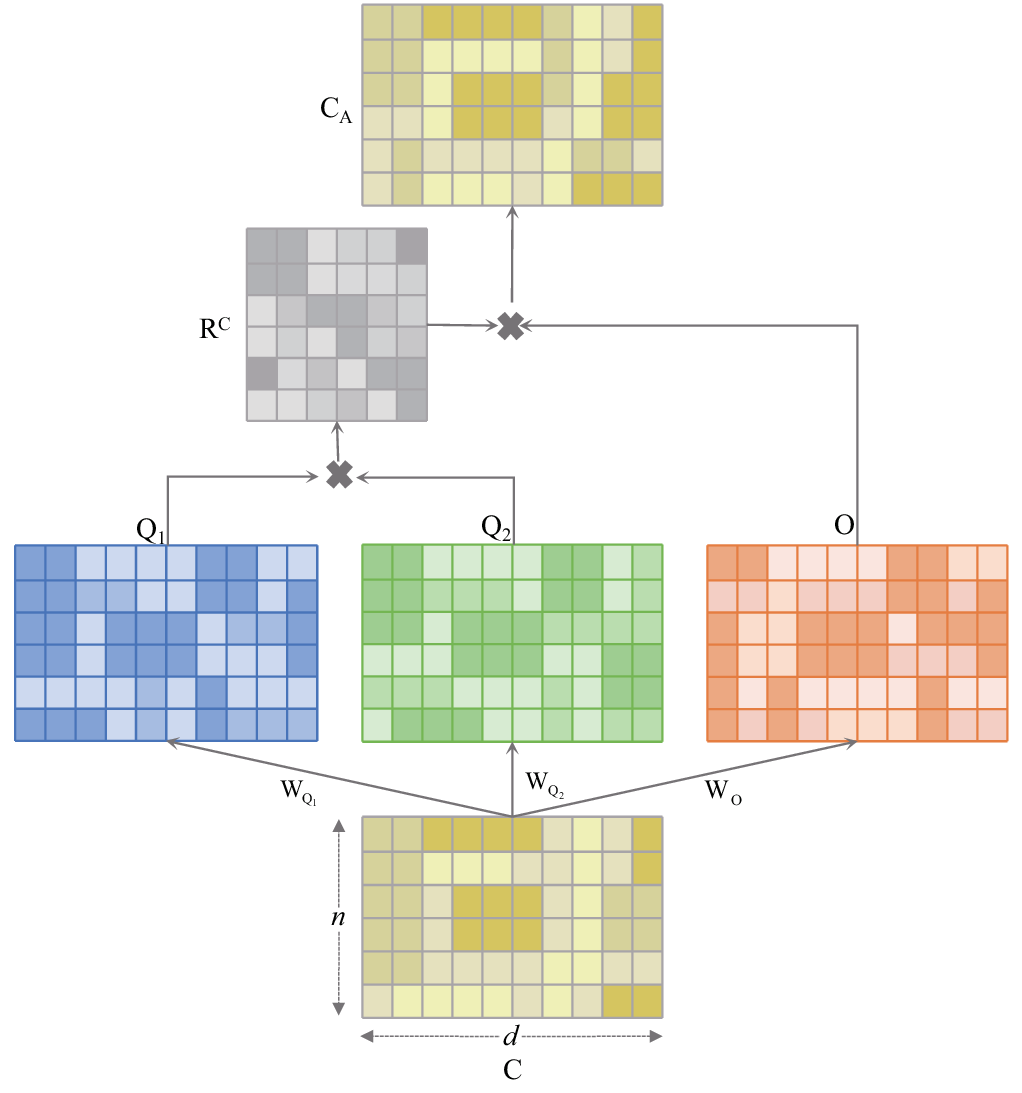}
		\caption{Attenion module. $\mathrm{W}_{\mathrm{Q}_{1}}$, $\mathrm{W}_{\mathrm{Q}_{2}}$, and $\mathrm{W_O}$ are utilized to achieve feature space transformation, and the weight matrix $\mathrm{R^C}$ is used to obtain instance pair weights.}
		\label{fig:attention}
	\end{figure}
	\subsubsection{Instance pair weights}
	Semantic features of instances can be utilized to establish the instance relationship because instances in the same cluster have the same semantic information. Inspired by \cite{yan2023gcfagg}, the attention mechanism of transformer \cite{vaswani2017attention} is employed to evaluate the relationships between instances, as shown in \cref{fig:attention}. We first concatenate all specific view semantic features $\mathrm{C}^{v}$ and the semantic feature $\hat{\mathrm{C}}$ of the fusion view to get $\mathrm{C}=\left[\mathrm{C}^{1},\mathrm{C}^{2},\dots,\mathrm{C}^{\mathrm{v}},\hat{\mathrm{C}}\right]$ , $\mathrm{C}\in\mathbb{R}^{\mathrm{n\times m}}$, where $\mathrm{m=(V+1)\times K}$. By including $\hat{\mathrm{C}}$ in the attention mechanism's input, we can extract more important information from the fusion view and lessen the influence of view-private information. C is mapped to different feature spaces by $\mathrm{W}_{\mathrm{Q}_{1}}$, $\mathrm{W}_{\mathrm{Q}_{2}}$ and $\mathrm{W_O}$.
	\begin{equation}
		\mathrm{Q}_1=\mathrm{C}\mathrm{W}_{\mathrm{Q}_1;}\mathrm{Q}_2=\mathrm{C}\mathrm{W}_{\mathrm{Q}_2;}\mathrm{O}=\mathrm{C}\mathrm{W_O}
	\end{equation}
	
	where $\mathrm{Q}_1\in\mathbb{R}^{\mathrm{n\times m}}$,$\mathrm{Q}_2\in\mathbb{R}^{\mathrm{n\times m}}$, $\mathrm{Q_O}\in\mathbb{R}^{\mathrm{n\times m}}$. We use the matrix $\mathrm{W}=\{\mathrm{W}_{\mathrm{Q}_{1}},\mathrm{W}_{\mathrm{Q}_{2}},\mathrm{W_O}\}$ as the parameters.
	
	The weight matrix $\mathrm{R^C}$ is defined as:
	\begin{equation}
		\mathrm{R^c=Softmax}{\left(\frac{\mathrm{Q_1}\mathrm{Q_2^T}}{\sqrt{m}}\right)}
		\label{eq:weight}
	\end{equation}
	
	The weight matrix $\mathrm{R^C}$ can represent the relationship between the instances, the higher the weight, the more similar the semantic features of the instance pairs and the higher the possibility of belonging to the same cluster. For negative instance pair $(\mathrm{i}, \mathrm{j})$, the instance pair weight is defined as: $1-\mathrm{R}_{\mathrm{ij}}^{\mathrm{C}}$.
	
	\subsubsection{Instance-level contrastive learning under semantic guidance}
	DCMCS apply instance-level contrastive learning to learn instance consistency across multiple views. Using a one-layer MLP, the instance-level features $\mathrm{H}^v=[\mathrm{h}_1^v,\mathrm{h}_2^v,\ldots,\mathrm{h}_{\mathrm{N}}^v]$, $\mathrm{H}^v\in\mathbb{R}^{\mathrm{N}\times\mathrm{d}_{\mathrm{h}}}$ of the specific view and the instance-level features $ \mathrm{\hat{H}}=[\mathrm{\hat{h}}_{1},\mathrm{\hat{h}}_{2},\ldots,\mathrm{\hat{h}}_{\mathrm{N}}]$,  $\widehat{\mathrm{H}}\in\mathbb{R}^{\mathrm{N\times d_h}}$ of the fusion view are acquired. $\mathrm{d_h}$ is the dimension of the instance-level features. We denote the parameter of feature MLP as
	$\mathrm{W}_\mathrm{H}$. The instances that belong to an individual sample in the fusion view and specific views are considered as positive pairs, while all other non-positive instances are treated as negative pairs, with the fusion view serving as the anchor.
	The influence of false negative pairs in contrastive learning is weakened by instance pair weights.
	
	Again, cosine similarity is utilized to measure the similarity between specific view instance-level features and fusion view instance-level features, defined as follows:
	\begin{equation}
	S(\hat{\mathrm{h}_i},\mathrm{h}_i^v)=\frac{\left\langle\hat{\mathrm{h}_i},\mathrm{h}_i^v\right\rangle}{\left\|\hat{\mathrm{h}_i}\right\|\left\|\mathrm{h}_i^v\right\|}
	\end{equation}
	The semantics-guided instance-level contrastive loss function is defined as:
	\begin{equation}
		\begin{aligned}&L_{i}=-\frac1{2N}\sum_{i=1}^N\sum_{v=1}^{V}\log\frac{e^{S(\hat{\mathrm{h}}_i,{\mathrm{h}}_i^v)/\tau_2}}{\sum_{j=1}^Ne^{(1-{\mathrm{R}}_{ij}^{{\mathrm{C}}})S(\hat{\mathrm{h}}_i,{\mathrm{h}}_j^v)/\tau_2}-e^{1/\tau_2}}\end{aligned}
	\end{equation}
	
	where $\tau_{2}$ denotes the temperature coefficient, and $1-\mathrm{R}_{ij}^{\mathrm{C}}$ denotes the instance pair weights. $\mathrm{R}_{ij}^{\mathrm{C}}$ is obtained from \cref{eq:weight}. A larger weight $\mathrm{R}_{ij}^{\mathrm{C}}$ indicates a higher likelihood of the instance pair belonging to the same cluster, the instance pair weight $1-\mathrm{R}_{ij}^{\mathrm{C}}$ is smaller. The instance pair weights can reduce the impact of false negative pairs and produce better cluster-friendly features.
	\subsection{Optimization}
	The final overall loss function is defined as follows:
	\begin{equation}
		L=L_{\mathrm{con}}+\lambda_1L_i+\lambda_2L_c
		\label{eq:loss}
	\end{equation}
	
	Among them $\lambda_{1}$ and $\lambda_{2}$ are the balance parameters. According to parameter experiment in \cref{sec:para}, we set $\lambda_{1}$ and $\lambda_{2}$ to 1.
	DCMCS obtain the cluster label $\mathrm{y}_i$ through the semantic features of the fusion view:
	\begin{equation}
		\mathrm{y}_i=\arg\max(\hat{\mathrm{c}}_i)
		\label{eq:result}
	\end{equation}

	The final clustering results $\mathrm{Y}=\left[y_{1},y_{2},...,y_\mathrm{N}\right]$ for each sample are obtained.
	A summary of our optimization process is shown in \cref{alg:algo}.
		\begin{center}
		\begin{minipage}{.7\linewidth}  
			\begin{algorithm}[H]
				\caption{our optimization process}
				\label{alg:algo}
				\begin{algorithmic}[1]
					\Require Multi-view dataset  $\{\mathrm{X}^v\}_{v=1}^\mathrm{V}$; Number of clusters K; Temperature parameters $\tau_{1}$ and $\tau_{2}$; Balance parameters $\lambda_{1}$ and $\lambda_{2}$.\\
					Initialize $\left\{\theta_{v}\text{,}{\eta_{v}}\right\}_{v=1}^{\mathrm{V}}$ by minimizing \cref{eq:con};\\
					Computing $\mathrm{\hat{Z}}$ via \cref{eq:fusion};\\
					Optimize $\mathrm{W}_\mathrm{H}$,$\mathrm{W}_\mathrm{C}$,$\left\{\theta_{v}\text{,}{\eta_{v}}\right\}_{v=1}^{\mathrm{V}}$ by minimizing \cref{eq:loss} \\
					Calculate semantic labels by \cref{eq:result}
					\Ensure  The label predictions $\mathrm{Y}=\left[y_{1},y_{2},...,y_\mathrm{N}\right]$.
				\end{algorithmic}
			\end{algorithm}
		\end{minipage}
\end{center}
	\section{Experiment}
	\label{sec:exper}
	\subsection{Experimental Setup}
	\begin{table}[t]
		\caption{The information of the datasets in our experiments}
		\label{tab:data}
		\centering
		\begin{tabular}{c|c|c|cc}
			\toprule
			Datasets     & Samples & Views & Classes   \\
			\midrule
			Synthetic3d   & 600     & 3     & 3         \\
			Hdigit  & 10000   & 2     & 10        \\
			Cifar10     & 50000   & 3     & 10        \\
			YouTube Face  & 101499  & 5     & 31       \\
			Caltech-5V   & 1400    & 5     & 7          \\
			\bottomrule
		\end{tabular}
	\end{table}
	\subsubsection{Datasets.}
	Information about the five public datasets used for the experiments is given in \cref{tab:data}. Synthetic3d \cite{kumar2011co} is a synthetic dataset containing 600 samples of 3 views. Hdigit\cite{chen2022representation} is a handwritten digit dataset obtained from MNIST Handwritten Digits and USPS Handwritten Digits. Cifar10\footnote{http://www.cs.toronto.edu/kriz/cifar.html}is a real RGB image dataset containing 10 clusters processed in the same way as in \cite{yan2023gcfagg}. YouTubeFace\footnote{https://www.cs.tau.ac.il/ wolf/ytfaces/} is a massive face dataset with five views that are sourced from the YouTube video database. We divide Caltech-5V\cite{fei2004learning} into multiple datasets with different numbers of views. Caltech-5V contains views WM, CENTRIST, LBP, GIST, and HOG, Caltech-2V contains the first two views, Caltech-3V contains the first three views and Caltech-4V contains the first four views.
	\begin{table}[t]
		\caption{Whether fusion, instance-level contrastive learning, or cluster-level contrastive learning are used in the methods.}
		\label{tab:method}
		\centering
		\begin{tabular}{c|c|c|c}
			\toprule
			Method         & Fusion & Instance-level& Cluster-level\\
			\midrule
			EAMC (2020)    & $\checkmark$      &                      &                      \\
			CONAN (2021)   & $\checkmark$      & $\checkmark$                    &                      \\
			SiMVC (2021)   & $\checkmark$      &                      &                     \\
			CoMVC (2021)   & $\checkmark$      & $\checkmark$                   &                     \\
			MFLVC (2022)   &        & $\checkmark$                   & $\checkmark$                  \\
			CVCL (2023)    &        &                      & $\checkmark$                   \\
			AECoDDC (2023) & $\checkmark$     & $\checkmark$                   &                     \\
			GCFAgg (2023)  & $\checkmark$      & $\checkmark$                    &                     \\
			\bottomrule
		\end{tabular}
	\end{table}
	\subsubsection{Comparison methods.} EAMC\cite{zhou2020end} uses adversarial learning to align features of different views and employs attention layer to fuse views. CONAN\cite{ke2021conan} proposes a deep fusion module and introduces intermediate variable to align view-specific features. SiMVC\cite{trosten2021reconsidering} analyses the problems of alignment and uses weighted fusion view features to obtain clustering results. CoMVC \cite{trosten2021reconsidering} builds on SiMVC by adding instance-level contrastive learning and designing it as a selection. MFLVC \cite{xu2022multi} splits the learning of common information and view-specific features into two distinct levels of space, performing contrastive learning at the high-level features. The cluster allocation results are further optimized for cluster-level contrastive learning using CVCL\cite{chen2023deepc}. AECoDDC is one of the approaches under the generalized framework proposed in \cite{trosten2023effects}. It aligns view features with contrastive learning. GCFAgg\cite{yan2023gcfagg} proposes a structure-guided instance-level contrastive loss to achieve consistency goal. \cref{tab:method} displays whether these methods make use of fusion, instance-level contrastive learning, or cluster-level contrastive learning.
	\subsubsection{Evaluation metrics.} The clustering effectiveness is evaluated by three metrics: clustering accuracy (ACC), normalized mutual information (NMI), and purity (PUR).
	
	\subsubsection{Implementation.}
	The experimental framework is implemented on the Pytorch platform and executed on a 24GB NVIDIA Geforce RTX 3090 Linux server. The encoder and decoder are implemented using fully connected layers with encoder dimensions of input-500-500-2000-512. The decoder is symmetric with encoder. Adam optimizer is adopted with a learning rate of 0.0003 and a batch size of 256. The experiments use the reconstruction loss to pre-train for 200 epochs, followed by 100 or 150 epochs using the overall loss function. The code will be released.
	\begin{table}[b]
		\centering
		\caption{Results of all methods on four datasets. Bold denotes the best results and underline denotes the second-best.}
		\label{tab:result1}
		\resizebox{1\textwidth}{0.7in}{

			\begin{tabular}{l|ccc|ccc|ccc|ccc}
				\toprule
				Datasets       &   
				\multicolumn{3}{c|}{Synthetic3d}                     & \multicolumn{3}{c|}{Hdigit}                          & \multicolumn{3}{c|}{Cifar10}                         & \multicolumn{3}{c}{YouTubuFace}                     \\
			\midrule
				Metrics        & ACC             & NMI             & PUR             & ACC             & NMI             & PUR             & ACC             & NMI             & PUR             & ACC             & NMI             & PUR             \\
				\midrule
				EAMC (2020)    & 0.9333          & 0.7688          & 0.9333          & 0.4878          & 0.5151          & 0.4939          & 0.4533          & 0.3824          & 0.4574          & 0.1366          & 0.0369          & 0.2662          \\
				CONAN (2021)   & 0.9650          & 0.8540          & 0.9650          & 0.9562          & 0.9193          & 0.9562          & 0.9255          & 0.8641          & 0.9255          & 0.1179          & 0.1178          & 0.1499          \\
				SiMVC (2021)   & 0.9366          & 0.7747          & 0.9366          & 0.7854          & 0.6705          & 0.7854          & 0.8359          & 0.7324          & 0.8359          & 0.0765          & 0.0481          & 0.2662          \\
				CoMVC (2021)   & 0.9530          & 0.8184          & 0.9520          & 0.9032          & 0.8713          & 0.9032          & 0.9275          & 0.8925          & 0.9275          & 0.1010          & 0.0851          & 0.2674          \\
				MFLVC (2022)   & 0.9650          & 0.8537          & 0.9650          & 0.9442          & 0.8750          & 0.9440          & 0.9918          & 0.9774          & 0.9918          & 0.2770          & 0.2952          & 0.3297          \\
				CVCL (2023)    & 0.6367          & 0.4188          & 0.6500          & 0.3216          & 0.2407          & 0.3227          & 0.9910          & 0.9755          & 0.9910          & 0.3116          & \underline{0.3431}    & 0.3840          \\
				AECoDDC (2023) &  \underline{0.9750}   &  \underline{0.8927}   &  \underline{0.9750}    &  \underline{0.9930}   &  \underline{0.9796}    &  \underline{0.9930}    & 0.8594          & 0.7590          & 0.8594          & 0.2703          & 0.2789          & 0.3578          \\
				GCFAgg (2023)  & 0.9700          & 0.8713          & 0.9700          & 0.9744          & 0.9305          & 0.9744          & \underline{0.9923}    &  \underline{0.9781}    &  \underline{0.9923}    &  \underline{0.3262}   & 0.3289          &  \underline{0.4007}    \\
				DCMCS(ours)           & \textbf{0.9817} & \textbf{0.9127} & \textbf{0.9817} & \textbf{0.9940} & \textbf{0.9811} & \textbf{0.9940} & \textbf{0.9929} & \textbf{0.9805} & \textbf{0.9929} & \textbf{0.3355} & \textbf{0.3508} & \textbf{0.4302} \\
			  \bottomrule
			\end{tabular}
		}
	\end{table}
	\begin{table}[t]
		\centering
		\caption{Results of all methods on Caltech with different views.  "-$X$V" indicates that there are $X$ views available.}
		\label{tab:result2}
		\resizebox{1\textwidth}{0.7in}{
			\begin{tabular}{l|ccc|ccc|ccc|ccc}
				\toprule
				Datasets       & \multicolumn{3}{c|}{Caltech-2V}                      & \multicolumn{3}{c|}{Caltech-3V}                      & \multicolumn{3}{c|}{Caltech-4V}                      & \multicolumn{3}{c}{Caltech-5V}                      \\
				 \midrule
				Metrics        & ACC             & NMI             & PUR             & ACC             & NMI             & PUR             & ACC             & NMI             & PUR             & ACC             & NMI             & PUR             \\
				 \midrule
				EAMC (2020)    & 0.4993          & 0.4449          & 0.5207          & 0.5014          & 0.3617          & 0.5029          & 0.4786          & 0.3709          & 0.4786          & 0.3936          & 0.3540          & 0.4000          \\
				CONAN (2021)   & 0.5750          & 0.4516          & 0.5757          & 0.5914          & 0.4981          & 0.5914          & 0.5571          & 0.5061          & 0.5735          & 0.7207          & 0.6418          & 0.7221          \\
				SiMVC (2021)   & 0.5083          & 0.4715          & 0.5573          & 0.5692          & 0.4953          & 0.5912          & 0.6193          & 0.5362          & 0.6303          & 0.7193          & 0.6771          & 0.7292          \\
				CoMVC (2021)   & 0.4663          & 0.4262          & 0.5272          & 0.5413          & 0.5043          & 0.5842          & 0.5683          & 0.5692          & 0.6463          & 0.7003          & 0.6871          & 0.7462          \\
				MFLVC (2022)   & 0.6060          & 0.5280          & 0.6160          & 0.6312          &  \underline{0.5663}    & 0.6392          & 0.7332          & 0.6523          & 0.7342          & 0.8042          & 0.7032          & 0.8043          \\
				CVCL (2023)    & 0.6479          &  \underline{0.5503}    & 0.6479          &  \underline{0.6664}    & 0.5493          & 0.6664          & 0.6643          & 0.5934          & 0.6864          & 0.7457          & 0.6549    & 0.5903          \\
				AECoDDC (2023) & 0.4579    &  0.3410   & 0.4607    &  0.5857    &  0.4549    &  0.6057   & 0.4893          & 0.3923          & 0.5271          & 0.6564          & 0.5898          & 0.6786          \\
				GCFAgg (2023)  &  \underline{0.6643}    & 0.5008          &  \underline{0.6643}    & 0.6400          & 0.5345          &  \underline{0.6529}    &  \underline{0.7343}    &  \underline{0.6610}    & \underline{0.7343}    &  \underline{0.8336}    &  \underline{0.7331}    &  \underline{0.8336}    \\
				DCMCS(ours)     & \textbf{0.6664} & \textbf{0.5709} & \textbf{0.6664} & \textbf{0.7543} & \textbf{0.6582} & \textbf{0.7600} & \textbf{0.8464} & \textbf{0.7291} & \textbf{0.8464} & \textbf{0.8907} & \textbf{0.8172} & \textbf{0.8907} \\
				 \bottomrule
		\end{tabular}}
	\end{table}
	\subsection{Performance Analysis}
	The experimental results on four datasets are shown in \cref{tab:result1}, from which we could have the following findings: (1) Our DCMCS achieves the best results on all metrics. DCMCS performs 2\% better on NMI than the best comparable method, AECoDDC, on the Synthetic3d dataset. This is because \textbf{we employ semantics-guided instance-level contrastive learning, which weakens the effect of false negative pairs.} (2)\cref{tab:method} presents the usage of instance-level contrastive learning and cluster-level contrastive learning in the comparison methods. As we account for both of them and introduce instance pair weights, our DCMCS performs better. \textbf{DCMCS combines instance-level contrastive learning with clustering goal, obtaining instance relationships through semantic features rather than view-specific features like GCFAgg}.(3) The comparison methods with and without fusion are also depicted in the \cref{tab:method}. 
	Each view feature includes common and meaningless view-private information\cite{xu2022multi}. The importance of each view is taken into consideration by the fusion view. However, the issue of view fusion-related representational deterioration can also affect the clustering results\cite{xu2024self}. \textbf{We use weighted fusion to alleviate the representation degradation problem and focus on the common information of the fused views to further alleviate the interference of view-private information.}
	
	We further analyze DCMCS's performance on datasets with varying numbers of views using the Caltech dataset. The results in \cref{tab:result2} indicate that (1) The efficiency of our approach gets better as the number of views rises. It proves that DCMCS is robust on datasets with different number of views. (2) Compared to the best method, GCFAgg, DCMCS increases ACC metric on the Caltech-4V dataset by around 13\%. The quality of the Caltech dataset varies depending on the view, as seen in \cite{xu2024self}. \textbf{The semantic features we adopted are more useful when there are significant variations across views.} (3) Compared to the methods MFLVC and CVCL which do not use fusion, our method improves ACC about 16\% and 11\% on Caltech-3V respectively, proving considering the importance of the different views is important.
	
\begin{figure}[t]
		\centering
		\begin{subfigure}{0.32\linewidth}
			\centering
			\includegraphics[height=2.5cm]{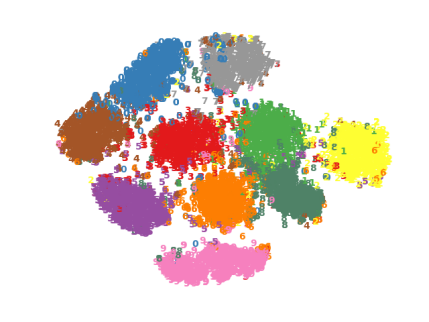}
			\caption{View 1}
			\label{fig:v1}
		\end{subfigure}
		\hfill
		\begin{subfigure}{0.32\linewidth}
				\includegraphics[height=2.5cm]{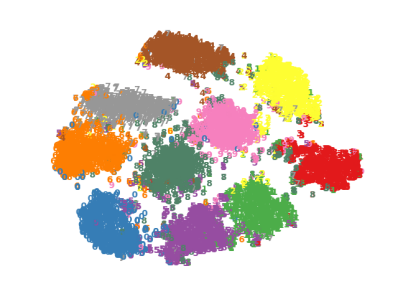}
			\caption{View 2}
			\label{fig:v2}
		\end{subfigure}
		\begin{subfigure}{0.32\linewidth}
			\includegraphics[height=2.5cm]{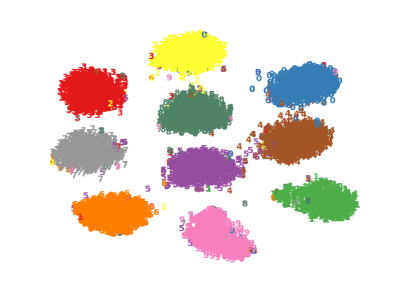}
			\caption{Fusion view}
			\label{fig:v3}
		\end{subfigure}
		
		\begin{subfigure}{0.32\linewidth}
			\centering
			\includegraphics[height=2.5cm]{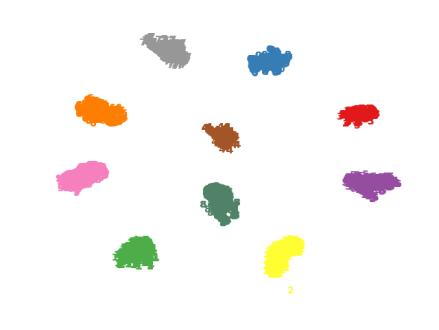}
			\caption{View 1}
			\label{fig:v4}
		\end{subfigure}
		\hfill
		\begin{subfigure}{0.32\linewidth}
			\includegraphics[height=2.5cm]{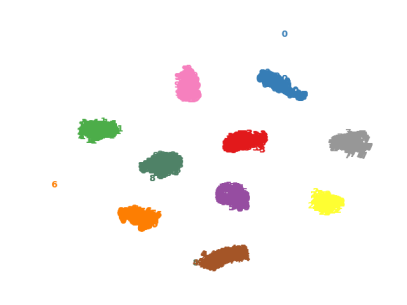}
			\caption{View 2}
			\label{fig:v5}
		\end{subfigure}
		\begin{subfigure}{0.32\linewidth}
			\includegraphics[height=2.5cm]{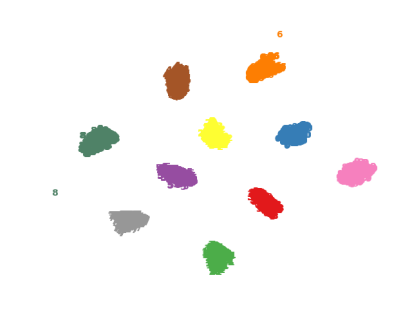}
			\caption{Fusion view}
			\label{fig:v6}
		\end{subfigure}
		\caption{
			Visualization of GCFAgg (a-c) and DCMCS (d-f) on Hdigit's two views and fusion view}
		\label{fig:visual}
	\end{figure}

	In \cref{fig:visual}, the visualization of GCFAgg \cref{fig:v1}-\cref{fig:v3} and our DCMCS \cref{fig:v4}-\cref{fig:v6} are achieved through the application of t-SNE on the two views of Hdigit and the fusion view. Our approach, in contrast to GCFAgg, enhances the clustering effect of views, resulting in a more compact intra-cluster structure and farther-separating inter-clusters. This is due to the fact that we introduce instance pair weights using semantic features and take into account cluster-level contrastive learning.

	\subsection{Ablation Studies}
	\label{sec:ablation}
	\begin{table}[t]
		\centering
		\caption{Ablation studies on loss components}
		\label{tab:com}
		\begin{tabular}{ccccllcclllll}
			\toprule
			\multicolumn{1}{c|}{}    & \multicolumn{3}{c|}{Components}                                                               & \multicolumn{2}{c|}{Caltech-5V}                           & \multicolumn{2}{c}{Hdigit}                                   \\ 
			 \midrule
			\multicolumn{1}{c|}{}    & \multicolumn{1}{c}{ ${L_\mathrm{con}}$}     & \multicolumn{1}{c}{ ${L}_{i}$}       & \multicolumn{1}{c|}{ ${L}_{c}$}       & \multicolumn{1}{c}{ACC}    & \multicolumn{1}{c|}{NMI}    & \multicolumn{1}{c}{ACC}      & \multicolumn{1}{c}{NMI}       \\ 
			 \midrule
			\multicolumn{1}{c|}{(a)} & \multicolumn{1}{c}{$\checkmark$}        & \multicolumn{1}{c}{}         & \multicolumn{1}{c|}{}         & \multicolumn{1}{c}{0.8850} & \multicolumn{1}{c|}{0.7957} & \multicolumn{1}{c}{0.9387}   & \multicolumn{1}{c}{0.8636} \\ 
			\multicolumn{1}{c|}{(b)} & \multicolumn{1}{c}{$\checkmark$}        & \multicolumn{1}{c}{$\checkmark$}        & \multicolumn{1}{c|}{}         & \multicolumn{1}{c}{0.4357} & \multicolumn{1}{c|}{0.4387} & \multicolumn{1}{c}{0.9525}   & \multicolumn{1}{c}{0.9091}   \\ 
			\multicolumn{1}{c|}{(c)} & \multicolumn{1}{c}{$\checkmark$}        & \multicolumn{1}{c}{}         & \multicolumn{1}{c|}{$\checkmark$}        & \multicolumn{1}{c}{0.7157} & \multicolumn{1}{c|}{0.6566} & \multicolumn{1}{c}{0.9919}   & \multicolumn{1}{c}{0.9753}   \\ 
			\multicolumn{1}{c|}{(d)} & \multicolumn{1}{c}{$\checkmark$}        & \multicolumn{1}{c}{$\checkmark$}        & \multicolumn{1}{c|}{$\checkmark$}        & \multicolumn{1}{l}{0.8907} & \multicolumn{1}{l|}{0.8172} & \multicolumn{1}{c}{0.9940}   & \multicolumn{1}{c}{0.9811}   \\ 
			\multicolumn{1}{c|}{(e)} & \multicolumn{1}{c}{}         & \multicolumn{1}{c}{$\checkmark$}        & \multicolumn{1}{c|}{$\checkmark$}        & \multicolumn{1}{l}{0.6657} & \multicolumn{1}{l|}{0.5661} & \multicolumn{1}{c}{0.9933}   & \multicolumn{1}{c}{0.9791}   \\ 
			\bottomrule
		\end{tabular}
	\end{table}
	\subsubsection{Loss components.}
	We use two datasets with distinct properties for the loss function ablation studies. Caltech-5V is a tiny dataset with varying view quality; Hdigit is large, and the two views' clustering accuracy is very little different, at 0.9420 and 0.9637. \cref{tab:com} displays the optimal results for both datasets that incorporate all loss functions. In case (a), we apply k-means\cite{bauckhage2015k} on the $\mathrm{\hat{Z}}$ to achieve clustering results. The findings show that improved features can be learned in the reconstruction. We use k-means on the $\mathrm{\hat{H}}$  in (b) to obtain clustering results. In comparison to (a), it improves on Hdigit but drastically degrades on Caltech-5V. When there is an adequate number of instances, instance-level contrastive learning is more helpful. In case (c), the Caltech-5V dataset effect is enhanced by about 39\% on ACC, which is consistent with the effect of\cite{chen2023deepc}. In addition, we attempt to remove the pre-training in (e), and the reduction is roughly 25\% on Caltech-5V and only 0.07\% on Hdigit. It suggests that in order to keep the model from crashing, pre-training is crucial on tiny datasets with varying view quality.
	\begin{table}[t]
		\centering
		\caption{Ablation studies with or without $\mathrm{R^C}$ on different datasets}
		\label{tab:rc1}
		\resizebox{1\textwidth}{0.4in}{
		\begin{tabular}{lccclllllllll}
		\toprule
			\multicolumn{1}{c|}{Datasets} & \multicolumn{3}{c|}{Synthetic3d}                                                        & \multicolumn{3}{c|}{Hdigit}                                                             & \multicolumn{3}{c|}{Cifar10}                                                            & \multicolumn{3}{c}{YouTubuFace}                                                        \\ \midrule
			\multicolumn{1}{l|}{Metrics}  & \multicolumn{1}{c}{ACC}    & \multicolumn{1}{c}{NMI}    & \multicolumn{1}{c|}{PUR}    & \multicolumn{1}{c}{ACC}    & \multicolumn{1}{c}{NMI}    & \multicolumn{1}{c|}{PUR}    & \multicolumn{1}{c}{ACC}    & \multicolumn{1}{c}{NMI}    & \multicolumn{1}{c|}{PUR}    & \multicolumn{1}{c}{ACC}    & \multicolumn{1}{c}{NMI}    & \multicolumn{1}{c}{PUR}    \\ \midrule
			\multicolumn{1}{l|}{w $\mathrm{R^C}$}     & \multicolumn{1}{c}{0.9817} & \multicolumn{1}{c}{0.9127} & \multicolumn{1}{c|}{0.9817} & \multicolumn{1}{c}{0.9940} & \multicolumn{1}{c}{0.9811} & \multicolumn{1}{c|}{0.9940} & \multicolumn{1}{c}{0.9929} & \multicolumn{1}{c}{0.9805} & \multicolumn{1}{c|}{0.9929} & \multicolumn{1}{c}{0.3355} & \multicolumn{1}{c}{0.3508} & \multicolumn{1}{c}{0.4302} \\
			\multicolumn{1}{l|}{w/o $\mathrm{R^C}$}   & \multicolumn{1}{c}{0.9800} & \multicolumn{1}{c}{0.9066} & \multicolumn{1}{c|}{0.9800} & \multicolumn{1}{c}{0.9923} & \multicolumn{1}{c}{0.9766} & \multicolumn{1}{c|}{0.9923} & \multicolumn{1}{c}{0.9916} & \multicolumn{1}{c}{0.9784} & \multicolumn{1}{c|}{0.9916} & \multicolumn{1}{c}{0.3037} & \multicolumn{1}{c}{0.3306} & \multicolumn{1}{c}{0.4059} \\    \bottomrule    
		\end{tabular}}
	\end{table}
	\begin{table}[t]
		\centering
		\caption{Ablation studies with or without $\mathrm{R^C}$ on Caletch}
		\label{tab:rc2}
		\resizebox{1\textwidth}{0.4in}{
		\begin{tabular}{l|lll|lll|lll|lll}
		\toprule
			Datasets & \multicolumn{3}{c|}{Caltech-2V}                                    & \multicolumn{3}{c|}{Caltech-3V}                                    & \multicolumn{3}{c|}{Caltech-4V}                                    & \multicolumn{3}{c}{Caltech-5V}                                    \\ \midrule
			Metrics  & \multicolumn{1}{c}{ACC}    & \multicolumn{1}{c}{NMI}    & \multicolumn{1}{c|}{PUR}    & \multicolumn{1}{c}{ACC}    & \multicolumn{1}{c}{NMI}    & PUR    & \multicolumn{1}{c}{ACC}    & \multicolumn{1}{c}{NMI}    & \multicolumn{1}{c|}{PUR}    & \multicolumn{1}{c}{ACC}    & \multicolumn{1}{c}{NMI}    & \multicolumn{1}{c} {PUR}    \\ \midrule
			w $\mathrm{R^C}$     & \multicolumn{1}{l}{0.6664} & \multicolumn{1}{l}{0.5709} & 0.6664 & \multicolumn{1}{l}{0.7543} & \multicolumn{1}{l}{0.6582} & 0.7600 & \multicolumn{1}{l}{0.8464} & \multicolumn{1}{l}{0.7291} & 0.8464 & \multicolumn{1}{l}{0.8907} & \multicolumn{1}{l}{0.8172} & 0.8907 \\
			w/o $\mathrm{R^C}$   & \multicolumn{1}{l}{0.5200} & \multicolumn{1}{l}{0.4262} & 0.5264 & \multicolumn{1}{l}{0.7064} & \multicolumn{1}{l}{0.6115} & 0.7350 & \multicolumn{1}{l}{0.7400} & \multicolumn{1}{l}{0.6762} & 0.7557 & \multicolumn{1}{l}{0.8314} & \multicolumn{1}{l}{0.7412} & 0.8314 \\ 
			\bottomrule
		\end{tabular}}
	\end{table}
	\subsubsection{Effectiveness of $\mathrm{R^C}$ .}The experimental results are shown in \cref{tab:rc1} and \cref{tab:rc2}. The effect is better with $\mathrm{R^C}$ than without it. The effect is improved by roughly 9.5\% on the ACC metric on the YouTubuFace dataset and somewhat on the Synthetic3d, Hdigit, and Cifar10 datasets. On datasets with large view quality disparities, Caltech-5V, the effect is significantly improved, with an average improvement of nearly 11\% on ACC. The influence of false negative pairs and the conflict between instance-level contrastive learning and the clustering target can be lessened when $\mathrm{R^C}$ is present. As a result, it is more suited for the clustering task, especially for views with significant quality differences.
	\begin{table}[H]
		\centering
		\caption{Ablation study on using $\mathrm{R^C}$ or $\mathrm{R^H}$, and  whether $\mathrm{\hat{C}}$ is added to $\mathrm{R^C}$ and $\mathrm{\hat{H}}$ is added to $\mathrm{R^H}$.}
		\label{tab:rcrh}
		\resizebox{1\textwidth}{0.5in}{
		\begin{tabular}{l|lll|lll|lll|lll}
			\hline
			Datasets  & \multicolumn{3}{c|}{Synthetic3d}                                   & \multicolumn{3}{c|}{Hdigit}                                        & \multicolumn{3}{c|}{Cifar10}                                       & \multicolumn{3}{c}{YouTubuFace}                                   \\ 
			 \midrule
			Metrics   & \multicolumn{1}{c}{ACC}    & \multicolumn{1}{c}{NMI}    & \multicolumn{1}{c|}{PUR}    & \multicolumn{1}{c}{ACC}    & \multicolumn{1}{c}{NMI}    & \multicolumn{1}{c|}{PUR}    & \multicolumn{1}{c}{ACC}    & \multicolumn{1}{c}{NMI}    & \multicolumn{1}{c|}{PUR}    & \multicolumn{1}{c}{ACC}    & \multicolumn{1}{c}{NMI}    & \multicolumn{1}{c}{PUR}    \\ 
			 \midrule
			$\mathrm{R^C}$ w $\mathrm{\hat{C}}$   & \multicolumn{1}{l}{0.9817} & \multicolumn{1}{l}{0.9127} & 0.9817 & \multicolumn{1}{l}{0.9940} & \multicolumn{1}{l}{0.9811} & 0.9940 & \multicolumn{1}{l}{0.9929} & \multicolumn{1}{l}{0.9805} & 0.9929 & \multicolumn{1}{l}{0.3355} & \multicolumn{1}{l}{0.3508} & 0.4302 \\ 
			$\mathrm{R^C}$ w/o $\mathrm{\hat{C}}$ & \multicolumn{1}{l}{0.9783} & \multicolumn{1}{l}{0.9028} & 0.9783 & \multicolumn{1}{l}{0.9933} & \multicolumn{1}{l}{0.9797} & 0.9933 & \multicolumn{1}{l}{0.9918} & \multicolumn{1}{l}{0.9776} & 0.9918 & \multicolumn{1}{l}{0.3299} & \multicolumn{1}{l}{0.3495} & 0.4230 \\
			$\mathrm{R^H}$ w $\mathrm{\hat{H}}$   & \multicolumn{1}{l}{0.9800} & \multicolumn{1}{l}{0.9078} & 0.9800 & \multicolumn{1}{l}{0.9937} & \multicolumn{1}{l}{0.9805} & 0.9937 & \multicolumn{1}{l}{0.9891} & \multicolumn{1}{l}{0.9718} & 0.9891 & \multicolumn{1}{l}{0.3024} & \multicolumn{1}{l}{0.3341} & 0.3846 \\
			$\mathrm{R^H}$ w/o $\mathrm{\hat{H}}$ & \multicolumn{1}{l}{0.9750} & \multicolumn{1}{l}{0.8899} & 0.9750 & \multicolumn{1}{l}{0.9925} & \multicolumn{1}{l}{0.9768} & 0.9925 & \multicolumn{1}{l}{0.9860} & \multicolumn{1}{l}{0.9644} & 0.9860 & \multicolumn{1}{l}{0.2931} & \multicolumn{1}{l}{0.3316} & 0.3770 \\ 
			 \bottomrule
		\end{tabular}}
	\end{table}
	\subsubsection{Comparison between $\mathrm{R^C}$ and $\mathrm{R^H}$}
	We analyze whether to employ semantic features to create the weight matrix $\mathrm{R^C}$ rather than instance-level features to produce $\mathrm{R^H}$, as well as whether to use the fusion view when generating $\mathrm{R^C}$ and $\mathrm{R^H}$. \cref{tab:rcrh} shows that the results obtained with $\mathrm{R^C}$ are better than those obtained with $\mathrm{R^H}$, with an improvement of 9.87\% on the YouTubuFace dataset on ACC. Because semantic features allow for a more accurate measurement of the relationship between instances. Furthermore, we note that joining $\mathrm{\hat{C}}$ in acquiring $\mathrm{R^C}$ and adding $\mathrm{\hat{H}}$ at the time of obtaining $\mathrm{R^H}$ both produce better results than not adding. This is because we focus on the common information of $\mathrm{\hat{C}}$, $\mathrm{\hat{H}}$ and alleviate the interference caused by view-private information of the fusion view.
	\begin{figure}[t]
		\centering
		\begin{subfigure}{0.26\linewidth}
			\centering
			\includegraphics[height=2.5cm]{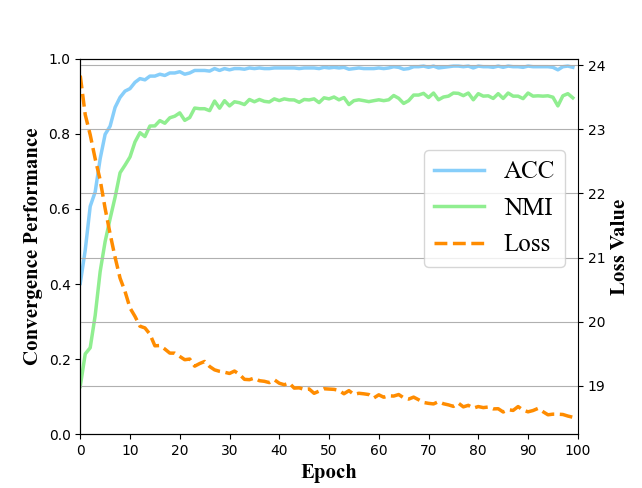}
			\caption{Loss vs. performance }
			\label{fig:loss}
		\end{subfigure}
		\hfill
		\begin{subfigure}{0.26\linewidth}
			\includegraphics[height=2.5cm]{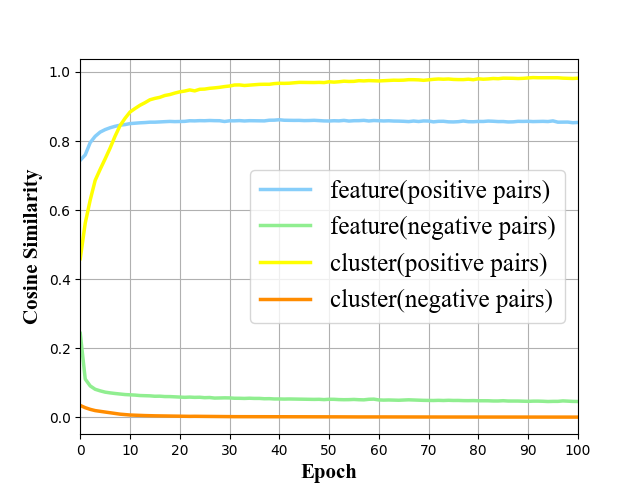}
			\caption{Similarities vs. CL}
			\label{fig:sim}
		\end{subfigure}
		\begin{subfigure}{0.22\linewidth}
			\centering
			\includegraphics[height=2.5cm]{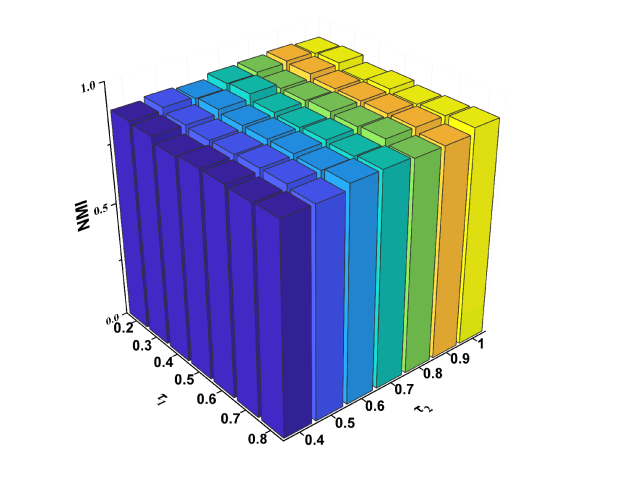}
			\caption{$\tau_{1}$ vs $\tau_{2}$ }
			\label{fig:tao}
		\end{subfigure}
		\begin{subfigure}{0.22\linewidth}
			\includegraphics[height=2.5cm]{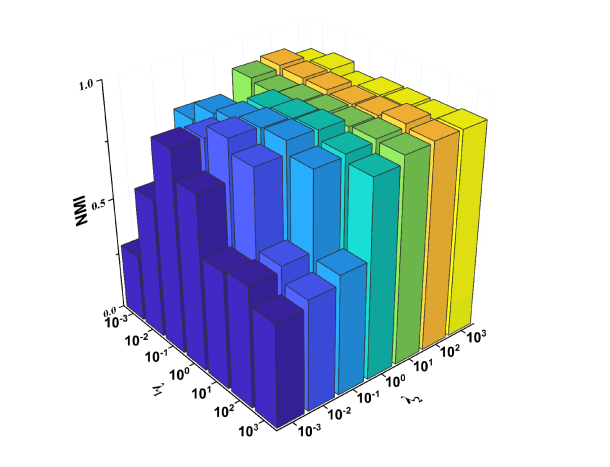}
			\caption{$\lambda_{1}$ vs. $\lambda_{2}$}
			\label{fig:la}
		\end{subfigure}
		\caption{(a) Convergence analysis. (b) The similarities of feature pairs and cluster pairs in contrative learning(CL). (c) and (d) Parameters sensitivity analysis.
		}
		\label{fig:conven}
	\end{figure}
	\subsection{Parameter Analysis.}
	\subsubsection{Convergence analysis.}
	\cref{fig:loss} illustrates how loss decreases as ACC and NMI increase until stability. DCMCS enjoys great convergence property. \cref{fig:sim} displays instances and clusters' positive and negative similarities. Instance-level features and cluster-level features show an increase in positive pairwise similarity and a drop in negative pairwise similarity after training. It matches our objective of clustering.

	\subsubsection{Parameter sensitivity analysis.}
	\label{sec:para}
	Through \cref{fig:tao}, we observe that the choice of temperature coefficients is insensitive for the semantics-guided instance-level contrastive loss and cluster-level contrastive loss. We set $\tau_{1}=1.0$ and $\tau_{2} = 0.5$ empirically. The overall loss function's hyper-parameters are denoted as $\lambda_{1}$ and $\lambda_{2}$. We can see from \cref{fig:la} that the model is insensitive when the $\lambda_{1}$ is at $10^{-1}\sim10^3$ and $\lambda_{2}$ is at $10^{0}\sim10^3$. We set $\lambda_{1}=1$ and $\lambda_{2}=1$ for convenience.
	\section{Conclusion}
	\label{sec:con}
	In this paper, we propose a semantics-guided contrastive multi-view clustering framework. Weighted fusion is used to fuse the features of views based on how important each view is. To lessen false negative pairs, instance pair weights are obtained from the semantic features of the fusion view and specific views through the attention mechanism. Furthermore, we minimize the influence of view-private information in the fusion view by concentrating on its common information. Experimental results on several public datasets demonstrate that DCMCS outperforms the state-of-the-art methods. Our framework's drawback is that it can be simple to assign the incorrect cluster when there are super-classes in the dataset, which leads to the instance being split up into a few clusters. We will address this issue in our future work. 

	%
	%
	\bibliographystyle{splncs04}
	\bibliography{main}

\begin{thebibliography}{10}
\providecommand{\url}[1]{\texttt{#1}}
\providecommand{\urlprefix}{URL }
\providecommand{\doi}[1]{https://doi.org/#1}

\bibitem{bauckhage2015k}
Bauckhage, C.: K-means clustering is matrix factorization. arXiv preprint
  arXiv:1512.07548  (2015)

\bibitem{chen2015total}
Chen, C., Li, X., Ng, M.K., Yuan, X.: Total variation based tensor
  decomposition for multi-dimensional data with time dimension. Numerical
  Linear Algebra with Applications  \textbf{22}(6),  999--1019 (2015)

\bibitem{chen2023multiview}
Chen, J., Mao, H., Peng, D., Zhang, C., Peng, X.: Multiview clustering by
  consensus spectral rotation fusion. IEEE Transactions on Image Processing
  (2023)

\bibitem{chen2023deepc}
Chen, J., Mao, H., Woo, W.L., Peng, X.: Deep multiview clustering by
  contrasting cluster assignments. arXiv preprint arXiv:2304.10769  (2023)

\bibitem{chen2022representation}
Chen, M.S., Lin, J.Q., Li, X.L., Liu, B.Y., Wang, C.D., Huang, D., Lai, J.H.:
  Representation learning in multi-view clustering: A literature review. Data
  Science and Engineering  \textbf{7}(3),  225--241 (2022)

\bibitem{chen2020simple}
Chen, T., Kornblith, S., Norouzi, M., Hinton, G.: A simple framework for
  contrastive learning of visual representations. In: International conference
  on machine learning. pp. 1597--1607. PMLR (2020)

\bibitem{chen2023deep}
Chen, W., Wang, H., Liang, C.: Deep multi-view contrastive learning for cancer
  subtype identification. Briefings in Bioinformatics  \textbf{24}(5),  bbad282
  (2023)

\bibitem{chen2023incomplete}
Chen, Z., Li, Y., Lou, K., Zhao, L.: Incomplete multi-view clustering with
  complete view guidance. IEEE Signal Processing Letters  (2023)

\bibitem{cui2023deep}
Cui, C., Ren, Y., Pu, J., Pu, X., He, L.: Deep multi-view subspace clustering
  with anchor graph. arXiv preprint arXiv:2305.06939  (2023)

\bibitem{fang2023comprehensive}
Fang, U., Li, M., Li, J., Gao, L., Jia, T., Zhang, Y.: A comprehensive survey
  on multi-view clustering. IEEE Transactions on Knowledge and Data Engineering
   (2023)

\bibitem{fei2004learning}
Fei-Fei, L., Fergus, R., Perona, P.: Learning generative visual models from few
  training examples: An incremental bayesian approach tested on 101 object
  categories. In: 2004 conference on computer vision and pattern recognition
  workshop. pp. 178--178. IEEE (2004)

\bibitem{guo2019anchors}
Guo, J., Ye, J.: Anchors bring ease: An embarrassingly simple approach to
  partial multi-view clustering. In: Proceedings of the AAAI conference on
  artificial intelligence. vol.~33, pp. 118--125 (2019)

\bibitem{guo2017improved}
Guo, X., Gao, L., Liu, X., Yin, J.: Improved deep embedded clustering with
  local structure preservation. In: Ijcai. vol.~17, pp. 1753--1759 (2017)

\bibitem{he2020momentum}
He, K., Fan, H., Wu, Y., Xie, S., Girshick, R.: Momentum contrast for
  unsupervised visual representation learning. In: Proceedings of the IEEE/CVF
  conference on computer vision and pattern recognition. pp. 9729--9738 (2020)

\bibitem{hu2023joint}
Hu, S., Zou, G., Zhang, C., Lou, Z., Geng, R., Ye, Y.: Joint contrastive
  triple-learning for deep multi-view clustering. Information Processing \&
  Management  \textbf{60}(3),  103284 (2023)

\bibitem{hu2017learning}
Hu, W., Miyato, T., Tokui, S., Matsumoto, E., Sugiyama, M.: Learning discrete
  representations via information maximizing self-augmented training. In:
  International conference on machine learning. pp. 1558--1567. PMLR (2017)

\bibitem{jin2021model}
Jin, Y., Li, C., Li, Y., Peng, P., Giannopoulos, G.A.: Model latent views with
  multi-center metric learning for vehicle re-identification. IEEE Transactions
  on Intelligent Transportation Systems  \textbf{22}(3),  1919--1931 (2021)

\bibitem{kang2020multi}
Kang, Z., Shi, G., Huang, S., Chen, W., Pu, X., Zhou, J.T., Xu, Z.: Multi-graph
  fusion for multi-view spectral clustering. Knowledge-Based Systems
  \textbf{189},  105102 (2020)

\bibitem{ke2021conan}
Ke, G., Hong, Z., Zeng, Z., Liu, Z., Sun, Y., Xie, Y.: Conan: contrastive
  fusion networks for multi-view clustering. In: 2021 IEEE International
  Conference on Big Data (Big Data). pp. 653--660. IEEE (2021)

\bibitem{ke2022mori}
Ke, G., Zhu, Y., Yu, Y.: Mori-ran: Multi-view robust representation learning
  via hybrid contrastive fusion. In: 2022 IEEE International Conference on Data
  Mining Workshops (ICDMW). pp. 467--474. IEEE (2022)

\bibitem{kumar2011co}
Kumar, A., Rai, P., Daume, H.: Co-regularized multi-view spectral clustering.
  Advances in neural information processing systems  \textbf{24} (2011)

\bibitem{lan2024double}
Lan, S., Zheng, Q., Yu, Y.: Double-level view-correlation multi-view subspace
  clustering. Knowledge-Based Systems  \textbf{284},  111271 (2024)

\bibitem{li2020bipartite}
Li, L., He, H.: Bipartite graph based multi-view clustering. IEEE transactions
  on knowledge and data engineering  \textbf{34}(7),  3111--3125 (2020)

\bibitem{li2023deep}
Li, W., Wang, S., Guo, X., Zhu, E.: Deep graph clustering with multi-level
  subspace fusion. Pattern Recognition  \textbf{134},  109077 (2023)

\bibitem{li2021contrastive}
Li, Y., Hu, P., Liu, Z., Peng, D., Zhou, J.T., Peng, X.: Contrastive
  clustering. In: Proceedings of the AAAI conference on artificial
  intelligence. vol.~35, pp. 8547--8555 (2021)

\bibitem{li2019deep}
Li, Z., Wang, Q., Tao, Z., Gao, Q., Yang, Z., et~al.: Deep adversarial
  multi-view clustering network. In: IJCAI. vol.~2, p.~4 (2019)

\bibitem{liang2020multi}
Liang, W., Zhou, S., Xiong, J., Liu, X., Wang, S., Zhu, E., Cai, Z., Xu, X.:
  Multi-view spectral clustering with high-order optimal neighborhood laplacian
  matrix. IEEE Transactions on Knowledge and Data Engineering  \textbf{34}(7),
  3418--3430 (2020)

\bibitem{liu2022inconsistency}
Liu, D., Peng, S.J., Liu, X., Zhu, L., Cui, Z., Li, T.: Inconsistency
  distillation for consistency: Enhancing multi-view clustering via mutual
  contrastive teacher-student leaning. In: 2022 IEEE International Conference
  on Data Mining (ICDM). pp. 251--258. IEEE (2022)

\bibitem{liu2013multi}
Liu, J., Wang, C., Gao, J., Han, J.: Multi-view clustering via joint
  nonnegative matrix factorization. In: Proceedings of the 2013 SIAM
  international conference on data mining. pp. 252--260. SIAM (2013)

\bibitem{long2023multi}
Long, Z., Zhu, C., Chen, J., Li, Z., Ren, Y., Liu, Y.: Multi-view mera subspace
  clustering. arXiv preprint arXiv:2305.09095  (2023)

\bibitem{peng2020deep}
Peng, X., Feng, J., Zhou, J.T., Lei, Y., Yan, S.: Deep subspace clustering.
  IEEE transactions on neural networks and learning systems  \textbf{31}(12),
  5509--5521 (2020)

\bibitem{ramon2018multi}
Ramon~Soria, P., Sukkar, F., Martens, W., Arrue, B.C., Fitch, R.: Multi-view
  probabilistic segmentation of pome fruit with a low-cost rgb-d camera. In:
  ROBOT 2017: Third Iberian Robotics Conference: Volume 2. pp. 320--331.
  Springer (2018)

\bibitem{tan2023sample}
Tan, Y., Liu, Y., Huang, S., Feng, W., Lv, J.: Sample-level multi-view graph
  clustering. In: Proceedings of the IEEE/CVF Conference on Computer Vision and
  Pattern Recognition. pp. 23966--23975 (2023)

\bibitem{trosten2021reconsidering}
Trosten, D.J., Lokse, S., Jenssen, R., Kampffmeyer, M.: Reconsidering
  representation alignment for multi-view clustering. In: Proceedings of the
  IEEE/CVF conference on computer vision and pattern recognition. pp.
  1255--1265 (2021)

\bibitem{trosten2023effects}
Trosten, D.J., L{\o}kse, S., Jenssen, R., Kampffmeyer, M.C.: On the effects of
  self-supervision and contrastive alignment in deep multi-view clustering. In:
  Proceedings of the IEEE/CVF Conference on Computer Vision and Pattern
  Recognition. pp. 23976--23985 (2023)

\bibitem{vaswani2017attention}
Vaswani, A., Shazeer, N., Parmar, N., Uszkoreit, J., Jones, L., Gomez, A.N.,
  Kaiser, {\L}., Polosukhin, I.: Attention is all you need. Advances in neural
  information processing systems  \textbf{30} (2017)

\bibitem{wang2021consistent}
Wang, Y., Chang, D., Fu, Z., Zhao, Y.: Consistent multiple graph embedding for
  multi-view clustering. IEEE transactions on multimedia  (2021)

\bibitem{wen2023highly}
Wen, J., Liu, C., Xu, G., Wu, Z., Huang, C., Fei, L., Xu, Y.: Highly confident
  local structure based consensus graph learning for incomplete multi-view
  clustering. In: Proceedings of the IEEE/CVF Conference on Computer Vision and
  Pattern Recognition. pp. 15712--15721 (2023)

\bibitem{xie2020adaptive}
Xie, D., Gao, Q., Wang, Q., Zhang, X., Gao, X.: Adaptive latent similarity
  learning for multi-view clustering. Neural Networks  \textbf{121},  409--418
  (2020)

\bibitem{xu2024self}
Xu, J., Chen, S., Ren, Y., Shi, X., Shen, H., Niu, G., Zhu, X.: Self-weighted
  contrastive learning among multiple views for mitigating representation
  degeneration. Advances in Neural Information Processing Systems  \textbf{36}
  (2024)

\bibitem{xu2022multi}
Xu, J., Tang, H., Ren, Y., Peng, L., Zhu, X., He, L.: Multi-level feature
  learning for contrastive multi-view clustering. In: Proceedings of the
  IEEE/CVF Conference on Computer Vision and Pattern Recognition. pp.
  16051--16060 (2022)

\bibitem{yan2023gcfagg}
Yan, W., Zhang, Y., Lv, C., Tang, C., Yue, G., Liao, L., Lin, W.: Gcfagg:
  Global and cross-view feature aggregation for multi-view clustering. In:
  Proceedings of the IEEE/CVF Conference on Computer Vision and Pattern
  Recognition. pp. 19863--19872 (2023)

\bibitem{yang2023one}
Yang, W., Wang, Y., Tang, C., Tong, H., Wei, A., Wu, X.: One step multi-view
  spectral clustering via joint adaptive graph learning and matrix
  factorization. Neurocomputing  \textbf{524},  95--105 (2023)

\bibitem{yang2023exploring}
Yang, X., Xu, W., Leng, D., Wen, Y., Wu, L., Li, R., Huang, J., Bo, X., He, S.:
  Exploring novel disease-disease associations based on multi-view fusion
  network. Computational and Structural Biotechnology Journal  \textbf{21},
  1807--1819 (2023)

\bibitem{yang2021deep}
Yang, X., Deng, C., Dang, Z., Tao, D.: Deep multiview collaborative clustering.
  IEEE Transactions on Neural Networks and Learning Systems  (2021)

\bibitem{yin2022incomplete}
Yin, J., Sun, S.: Incomplete multi-view clustering with cosine similarity.
  Pattern Recognition  \textbf{123},  108371 (2022)

\bibitem{yu2018web}
Yu, H., Zhang, T., Chen, J., Guo, C., Lian, Y.: Web items recommendation based
  on multi-view clustering. In: 2018 IEEE 42nd annual computer software and
  applications conference (COMPSAC). vol.~1, pp. 420--425. IEEE (2018)

\bibitem{zhang2023multi}
Zhang, L., Lin, L., Li, J.: Multi-view clustering by cps-merge analysis with
  application to multimodal single-cell data. PLOS Computational Biology
  \textbf{19}(4),  e1011044 (2023)

\bibitem{zhong2023self}
Zhong, G., Pun, C.M.: Self-taught multi-view spectral clustering. Pattern
  Recognition  \textbf{138},  109349 (2023)

\bibitem{zhou2020end}
Zhou, R., Shen, Y.D.: End-to-end adversarial-attention network for multi-modal
  clustering. In: Proceedings of the IEEE/CVF conference on computer vision and
  pattern recognition. pp. 14619--14628 (2020)

\bibitem{zhou2024mcoco}
Zhou, Y., Zheng, Q., Wang, Y., Yan, W., Shi, P., Zhu, J.: Mcoco: Multi-level
  consistency collaborative multi-view clustering. Expert Systems with
  Applications  \textbf{238},  121976 (2024)

\bibitem{zong2018multi}
Zong, L., Zhang, X., Liu, X.: Multi-view clustering on unmapped data via
  constrained non-negative matrix factorization. Neural Networks  \textbf{108},
   155--171 (2018)

\end{thebibliography}
\end{document}